\documentclass[sigconf,nonacm]{acmart}

\usepackage{booktabs}
\usepackage{amsmath}
\usepackage{multirow}
\usepackage{graphicx}
\usepackage{array}

\newcommand{\leadin}[1]{\par\medskip\noindent\textit{#1}}

\acmConference[KDD '22]{Proceedings of the 28th ACM SIGKDD Conference on
  Knowledge Discovery and Data Mining}{August 14--18, 2022}{Washington, DC, USA}
\acmDOI{10.1145/3534678.3539069}

\begin{document}

\title{COBART: Controlled, Optimized, Bidirectional and Auto-Regressive
  Transformer for Ad Headline Generation}
\titlenote{\copyright{} Kanungo, Das, Pooja A, and Negi 2022. This is the
  author's version of the work. It is posted here for your personal use. Not for
  redistribution. The definitive Version of Record was published in
  \textit{Proceedings of the 28th ACM SIGKDD Conference on Knowledge Discovery
  and Data Mining (KDD '22), August 14--18, 2022, Washington, DC, USA},
  \url{https://doi.org/10.1145/3534678.3539069}.}

\author{Yashal Shakti Kanungo}
\authornote{Equal contribution.}
\affiliation{%
  \institution{Amazon}
  \country{United States of America}
}

\author{Gyanendra Das}
\authornotemark[1]
\authornote{Work done during internship at Amazon.}
\affiliation{%
  \institution{Amazon}
  \country{India}
}

\author{Pooja A}
\affiliation{%
  \institution{Amazon}
  \country{India}
}

\author{Sumit Negi}
\affiliation{%
  \institution{Amazon}
  \country{India}
}

\renewcommand{\shortauthors}{Yashal Kanungo et al.}

\begin{abstract}
Online ads are essential to all businesses and ad headlines are one of their
core creative component. Existing methods can generate headlines automatically
and also optimize their click-through-rate (CTR) and quality. However, evolving
ad formats and changing creative requirements make it difficult to generate
optimized \& customized headlines. We propose a novel method that uses prefix
control tokens along with BART~\cite{lewis2019bart} fine-tuning. It yields the
highest CTR and also allows users to control the length of generated headlines
for use across different ad formats. The method is also flexible and can easily
be adapted to other architectures, creative requirements and optimization
criteria. Our experiments demonstrate a 25.82\% increment in Rouge-L and a
5.82\% increment in estimated CTR over previously published strong ad headline
generation baseline.
\end{abstract}


\keywords{ad generation; controlled generation; sponsored advertising; ad
  optimization; headline generation; E-commerce}

\maketitle

\begin{figure*}[t]
  \centering
  \includegraphics[width=\textwidth]{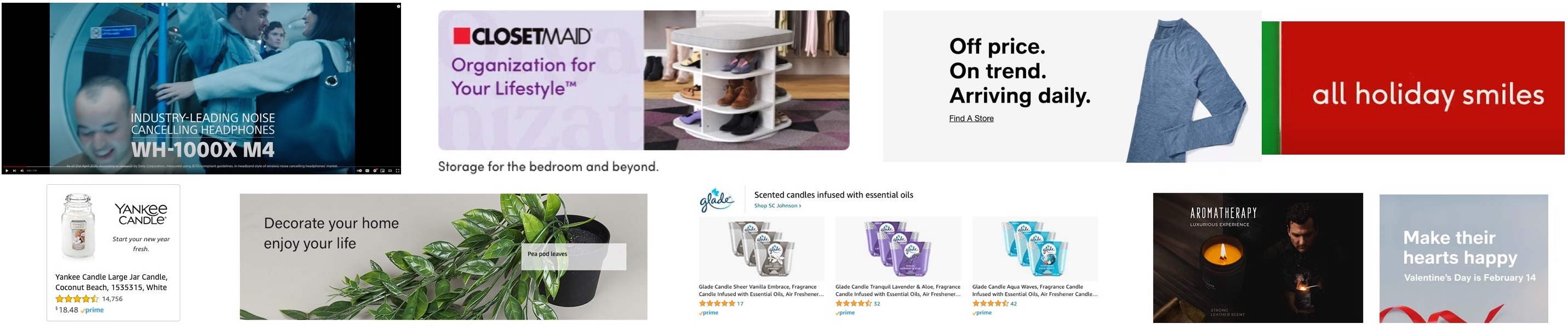}
  \caption{Ads from across the internet with different headlines. These include
    text-only ads, image+text ads, or screengrabs from video ads with headlines
    included within the video. The headlines have varying lengths, messaging, and
    advertising intent. Some headlines are informative with no adjectives, some
    are short, and some others tend to be much longer.}
  \label{fig:ads}
\end{figure*}

\begin{table}[t]
  \small
  \setlength{\tabcolsep}{4pt}
  \newcommand{\hc}[1]{\begin{tabular}[t]{@{}c@{}}#1\end{tabular}}
  \begin{tabular}{@{}>{\raggedright\arraybackslash}p{1.35cm}
                     >{\centering\arraybackslash}p{1.5cm}
                     >{\centering\arraybackslash}p{1.65cm}
                     >{\centering\arraybackslash}p{1.8cm}@{}}
    \toprule
    \multicolumn{4}{@{}l}{\textbf{Input} (Single ad with multiple products; two shown)}\\
    \midrule
    \textbf{\mbox{Product 1}} & \multicolumn{3}{@{}p{5.3cm}@{}}{Boyd's Cosmetics NYC- No Lines Temporary Wrinkle Remover with our Brush It Away Medium \ldots}\\
    \textbf{\mbox{Product 2}} & \multicolumn{3}{@{}p{5.3cm}@{}}{No Lines Temporary Wrinkle Remover \ldots}\\
    \midrule
    \multicolumn{4}{c}{\textbf{Generated Headlines}}\\
    \midrule
    \textbf{UniLM} & \multicolumn{3}{c}{\textit{Boyd's New York City Cosmetics}}\\
    \addlinespace
    \textbf{BART} & \multicolumn{3}{c}{\textit{New \& improved anti-aging with our brush}}\\
    \midrule
    \textbf{COBART} & \multicolumn{3}{c}{\textbf{Length} $\rightarrow$}\\
    \textbf{CTR} $\downarrow$ & Short & Medium & Long\\
    \midrule
    \hc{Low\\(Bucket\\1/15)} & \textit{Boyd's Cosmetics NYC- No Lines} & \textit{Save Now on New York City Collections} & \textit{No Lines Temporary Wrinkle Remover with Brush It Away}\\
    \addlinespace
    \hc{Medium\\(Bucket\\7/15)} & \textit{Shop Boyd's NYC, No Lines Remover} & \textit{Tackle Fine Lines \& Wrinkles with Boyd's NYC} & \textit{Boyd's Cosmetics NYC - No Lines Temporary Wrinkle Remover}\\
    \addlinespace
    \hc{High\\(Bucket\\15/15)} & \textit{Shop New York City's \#1 Makeup Brand} & \textit{New \& improved anti-aging with our brush - Boyd's NYC} & \textit{A New Generation of Makeup That Targets Fine Lines with a New Look}\\
    \bottomrule
  \end{tabular}
  \caption{Ad headline generation for a multi-product ad. Our method allows to
    \textit{control} and \textit{optimize} the length and click-through-rate
    (CTR) of headlines during inference.}
  \label{tab:teaser}
\end{table}

\section{Introduction}
Online advertising is an evolving domain that is now integral to all businesses.
Organizations across the world use ads to improve discovery and reach of their
businesses. Sellers on E-commerce platforms use online ads to build brand
awareness and connect shoppers to their products.

One of the key component of ads is their `headline'. Headlines are a part of all
ad formats including text, image, and video ads as illustrated in
Figure~\ref{fig:ads}. They summarize the key properties of the underlying
products and promote the customers to engage. They may also be the first preview
of the product/brand for shoppers. Thus having the right style and appeal is
important to attract shopper attention. Style and content are not the only
factors for generating headlines. Length is equally important for ads across
different formats and target device screen sizes. Long headlines are used in
text ads, medium length headlines are shown alongside images and shorter
headlines are typically overlaid on images and videos. Mobile-only ad formats
also encourage text ads with shorter headlines.

Language barriers, varying shopper interests, market trends, seasonal trends,
varying inventory of products, changing style preferences etc. add to the
complexity and scale of the problem. With rapidly evolving product catalogs with
sometimes billions of products, manually writing such headlines is infeasible
and will require extensive human effort. The content present within the product
catalog may also have a high variance in quality. This makes it difficult to use
it as reference, leading to keyword stuffing and poor headlines. It is thus
difficult to write these headlines for different products, ad formats and
requirements at scale.

It is not sufficient to just automate the headline generation process. One of
the key metrics that is used to measure headline attractiveness and engagement
is the click-through-rate (CTR). It is the ratio of number of clicks a headline
receives to the number of times the headline was shown. Thus, ideal headlines
need to be automatically written, meet platform and policy requirements, meet
creative requirements (length, style etc.) and yield high CTR. Moreover, any
automatic solution that solves this problem practically needs to be highly
scalable and extendable to meet a wide range of requirements.

We propose a novel method to fine-tune a pre-trained Bidirectional and
Auto-Regressive Transformer (BART)~\cite{lewis2019bart} model to generate ad
headlines for multiple input products. It jointly learns to control and optimize
any desired characteristic of the headline such as CTR and length using special
control tokens. These control tokens are provided as input to the bidirectional
encoder and allow the encoder and decoder's Attention to be conditioned to
control CTR and length. We compare our proposed method to other state-of-the-art
controlled generation methods~\cite{hu2018toward}, other ad headline generation
method~\cite{kanungo2021ad}, and multiple other baselines and ablations. An
example ad with headlines generated by baselines and our method is illustrated
in the Table~\ref{tab:teaser}.

Our key contributions are:
\begin{itemize}
  \item Our method allows users to control and optimize desired characteristics
    of generation at inference time.
  \item Our method allows to mix-and-match multiple characteristics. We also show
    that it can be combined with other optimization techniques to further improve
    the performance.
  \item To validate our proposal, we build extensive baselines, try multiple
    variations \& ablations, and demonstrate significant improvement in Rouge-L
    and CTR. The results show that it is possible to effectively use control
    tokens to fine-tune language models that were pre-trained without them.
  \item Thus, our proposal solves a large-scale real-world problem using a novel
    method with multiple practical applications. It also does not have any
    negative impact on model inference latency and can replace existing methods
    with minimal changes.
\end{itemize}

\section{Related Work}
\leadin{Natural Language Generation.} Transformer~\cite{vaswani2017attention}
based methods for Natural Language Generation (NLG) have shown great potential in
the recent years. They span all possible combinations of the Transformer encoder
and decoder, including encoder-only NLG~\cite{dong2019unified}, decoder-only
NLG~\cite{brown2020language}, encoder-decoder
NLG~\cite{lewis2019bart,qi2020prophetnet,raffel2020exploring,song2019mass},
combining differently pre-trained models~\cite{rothe2020leveraging}, and joint
pre-training~\cite{song2019mass}. Methods~\cite{aribandi2021ext5,raffel2020exploring}
have shown that it is possible to use task specific prefixes to train a single
model for multiple tasks. All of these methods outperform earlier LSTM +
Attention~\cite{see2017get} based methods by implicitly generating better
headlines. These methods can be used for headline generation but they do not
perform any additional optimization to improve CTR or control the generation.

\leadin{Controlled NLG.} In~\cite{hu2018toward}, authors propose to use LSTM
based VAEs in a generator-discriminator setting to condition the generator output
on the VAE and an additional semantic style code to control the generation.
However, they target each feature to be controlled independently.
CTRL~\cite{keskar2019ctrl} proposes to condition the output of a decoder Language
Model (LM) on specific control codes in order to augment the generation. Decoder
only models offer lesser flexibility for conditional text generation compared to
encoder-decoder models such as~\cite{lewis2019bart}. PPLM~\cite{dathathri2020plug}
combines Transformer based LM with bag-of-words from different topics of choice
to change style. FUDGE~\cite{yang2021fudge} models the style conditional LMs using
Bayesian factorization instead of modeling the conditional generation directly.
MuCoCo~\cite{kumar2021controlled} combines pretrained LMs with differentiable
constraints to frame the generation as a constrained optimization problem and
control the generation. However, it is relatively slow for practical deployment.

\leadin{Headline Generation.} Headline Generation is akin to other text-to-text
generation tasks such as summarization. The discussed NLG methods can be applied
for headline generation, and various methods have proposed the use of Universal
Transformers~\cite{gavrilov2019self}, encoding structure of the text for better
headlines~\cite{zhang2020structure}, or using Reinforcement Learning (RL) based
techniques along with Transformer based architecture~\cite{kanungo2021ad} for
improving the quality of generation. As discussed earlier, the attractiveness of
headlines may be judged by the CTR they yield. Optimized headlines yield higher
CTR. Reinforcement Learning (RL) based techniques such as Actor-Critic,
Self-critical Sequence Training (SCST)~\cite{rennie2017self} etc. have been shown
to improve ad headline generation CTR when used with
LSTMs~\cite{hughes2019generating,song2020attractive,xu2019clickbait}.
\cite{jin2020hooks} proposes to use style dependent Layer Norm and
Transformer-Query transformation to transform headlines. However, these methods
either implicitly improve generation quality \& do not allow for explicit control
during inference or do not take advantage of in-domain datapoints and observed
CTR.

In our method, we extend the BART encoder-decoder
architecture~\cite{lewis2019bart} with characteristic control tokens (COBART:
Section~\ref{sec:cobart}). To evaluate our method, we also create two more
alternative extensions to BART that optimize CTR. The first (SCBART:
Section~\ref{sec:scbart}) uses Self-critical training~\cite{rennie2017self}. The
second extension (VBART: Section~\ref{sec:vbart}) uses variational Transformer
encoder with a generator and a discriminator~\cite{hu2018toward}.

\section{Methods}
Every ad consists of a set of one or more products $P$ such that the $i^{th}$
product is represented by the tokens in its title
$x_i = (x^i_1, x^i_2 \ldots x^i_{|x_i|})$.

During training, we have access to the original ad headline tokens
$H = (h_1, h_2 \ldots h_{|h|})$. We also have a set of observed or computed
characteristics of each headline $\Phi = \{\phi_1, \phi_2 \ldots \phi_{|\Phi|}\}$.
Examples of these characteristics include the observed continuous CTR value, the
discreet number of words present in the headline, categorical season in which the
headline was advertised, binary variables such as presence of brand names etc.

We primarily experiment with controlling and optimizing the observed CTR of the
headlines ($\phi_{ctr}$) and also include results of using the number of words
present in the headline ($\phi_{Length}$) without any loss of generality. These
two aspects define the control over attractiveness and semantics ($\phi_{ctr}$)
and the structure of the headlines ($\phi_{Length}$).

During inference, we are required to generate the headline
$\hat{H} = (\hat{h}_1, \hat{h}_2 \ldots \hat{h}_{|\hat{h}|})$ given the input
products $P$ and the desired level of attractiveness and length.

\subsection{Controlled and Optimized BART}
\label{sec:cobart}
The Bidirectional and Auto-Regressive Transformer (BART) model uses the
Transformer~\cite{vaswani2017attention} architecture as a denoising autoencoder
for pretraining. It is pretrained using a novel text infilling scheme combined
with sentence permutation. For an original sentence ($x_1 x_2 x_3 . x_4 x_5 .$)
BART encodes ($x_4 x_5 . x_1 \_ .$) using a bi-directional Transformer encoder.
It then passes the encoder's output to an auto-regressive left-to-right
Transformer decoder that regenerates the original sentence
($x_1 x_2 x_3 . x_4 x_5 .$) by infilling ($x_2, x_3$) and permuting the sentences
to the correct order.

\leadin{Headline generation without any controllable characteristics.} We
fine-tune a pretrained BART model~\cite{lewis2019bart} following the input
strategy used in~\cite{kanungo2021ad}. We input the product title tokens
($x^1_1, x^1_2 \ldots x^1_{|x_1|}$), with titles from different products
concatenated using an otherwise unused separator token. An example input would
be:
\begin{equation}
  X = (x^1_1, x^1_2, [SEP], x^2_1, [SEP], x^3_1)
\end{equation}

We thus minimize the loss:
\begin{equation}
  \mathcal{L}_{BART} = - \log \prod_{t=1}^{|H|} p(h_t | h_{1:t-1}, X)
\end{equation}

\begin{figure}[t]
  \centering
  \includegraphics[width=\columnwidth]{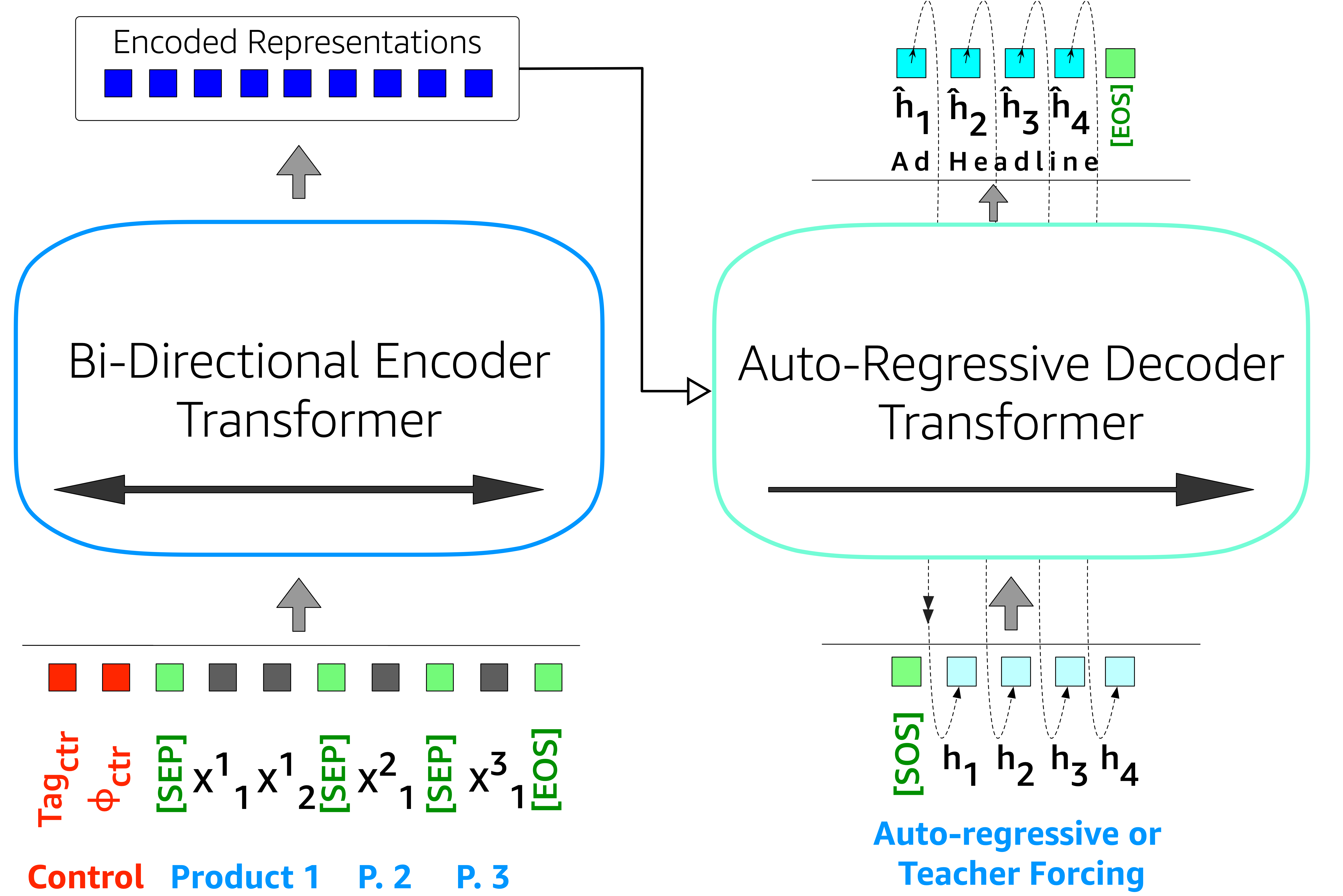}
  \caption{The COBART model takes computed characteristic control tokens ($\Phi$)
    based prefixes as input to the bi-directional encoder during training. During
    inference, these are replaced by user desired characteristics and passed on to
    the decoder for controlled and optimized generation.}
  \label{fig:cobart}
\end{figure}

\leadin{Headline generation with controllable characteristics.} We propose a
simple yet highly effective addition to jointly learn headline generation along
with optimization of characteristics. We propose to add a control token as prefix
to the input for each characteristic $\Phi = \{\phi_1, \phi_2 \ldots \phi_{|\Phi|}\}$
that we want to optimize and control. We compute the characteristics $\Phi$ for
each headline in the training set and add them as multiple prefixes along with
corresponding tags. Figure~\ref{fig:cobart} illustrates the model. For instance,
to optimize headline CTR, we bucketize the observed CTR $\phi_{ctr}$ for each
headline into multiple categorical buckets and then use it as an additional
control token in the input. We perform all our experiments with $\phi_{ctr}$ and
also include results for experiments to control the length of the generated
headline using the control token $\phi_{Length}$.

An example input for COBART would be:
\begin{equation}
  X_{CO} = (Tag_{ctr}, \phi_{ctr}, [SEP], x^1_1, x^1_2, [SEP], x^2_1, [SEP], x^3_1)
\end{equation}

This can be extended and combined with control tokens for any desired
characteristic present within the training set. The Attention formulation
automatically allows the encoded representation of the input to be updated based
on the control tokens without any modifications to the Transformer architecture.

\subsection{Self-critical BART}
\label{sec:scbart}
\begin{figure}[t]
  \centering
  \includegraphics[width=\columnwidth]{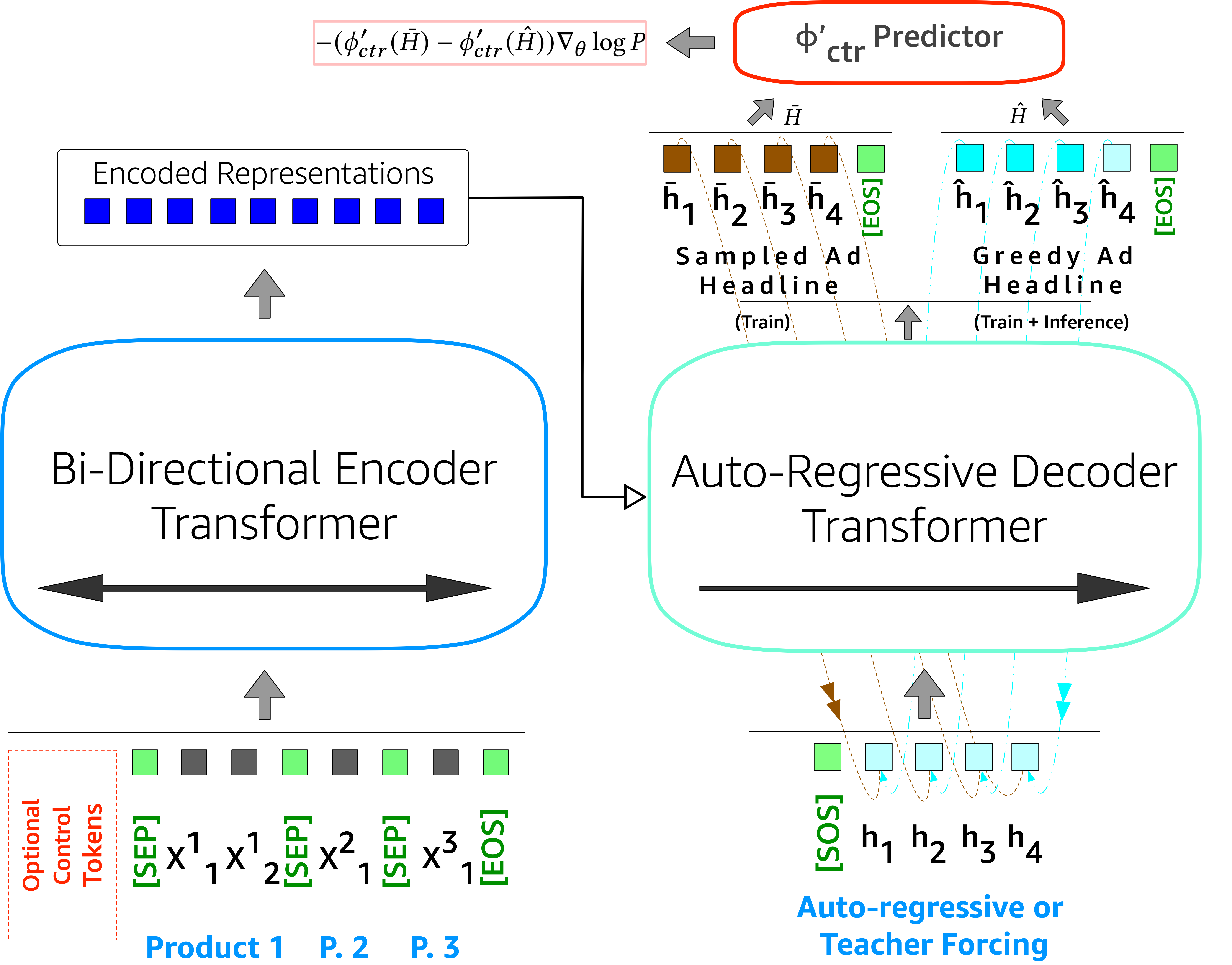}
  \caption{The Self-critical BART uses RL based reward difference between sampled
    and greedy headlines generated from the decoder to update the model. We also
    optionally use control tokens to obtain Self-critical COBART.}
  \label{fig:scbart}
\end{figure}

We extend previous Self-critical Sequence Training (SCST)~\cite{rennie2017self} ad
headline generation methods~\cite{hughes2019generating,kanungo2021ad} by
experimenting the usage of SCST for improving CTR of headlines generated by a
pretrained and fine-tuned BART model. In this method illustrated in
Figure~\ref{fig:scbart}, instead of using the original observed CTR $\phi_{ctr}$
of train headlines, we use an estimate $\phi'_{ctr}$ of each generated headline
as a reward that drives the optimization criterion. The estimate $\phi'_{ctr}$ is
predicted by an oracle model based on the DeBERTa~\cite{he2021deberta}
architecture.

For parameters $\theta$ of the BART model, we follow the policy $\pi_\theta$ and
generate the headline $\bar{H}$ by sampling. We then aim to minimize the negative
expected reward $\phi'_{ctr}(\bar{H})$.
\begin{equation}
  \mathcal{L}_{RL} = -\mathbb{E}_{\bar{H} \sim \pi_\theta}[\phi'_{ctr}(\bar{H})]
\end{equation}

Using the REINFORCE trick~\cite{williams1992simple} and SCST, we can estimate the
gradient to optimize $\phi'_{ctr}$ as:
\begin{equation}
  \nabla_\theta \mathcal{L}_{SC\_BART} \approx
    -(\phi'_{ctr}(\bar{H}) - \phi'_{ctr}(\hat{H})) \nabla_\theta \log P_{sc}
\end{equation}
where $\hat{H}$ is the headline generated by BART using greedy strategy, X is the
input from Eq 1 and
\begin{equation}
  P_{sc} = \prod_{t=1}^{|\bar{H}|} p(h_t | h_{1:t-1}, X)
\end{equation}
We train the model using the following combined loss with hyperparameter
$\lambda$:
\begin{equation}
  \mathcal{L}_{SC\_Total} = \lambda * \mathcal{L}_{BART} + (1-\lambda) * \mathcal{L}_{SC\_BART}
\end{equation}
The proposed usage of characteristics control tokens is directly compatible with
the Self-critical framework. Thus, we also experiment with Self-critical COBART
which combines SCST with COBART.

\subsection{Variational BART}
\label{sec:vbart}
\begin{figure}[t]
  \centering
  \includegraphics[width=\columnwidth]{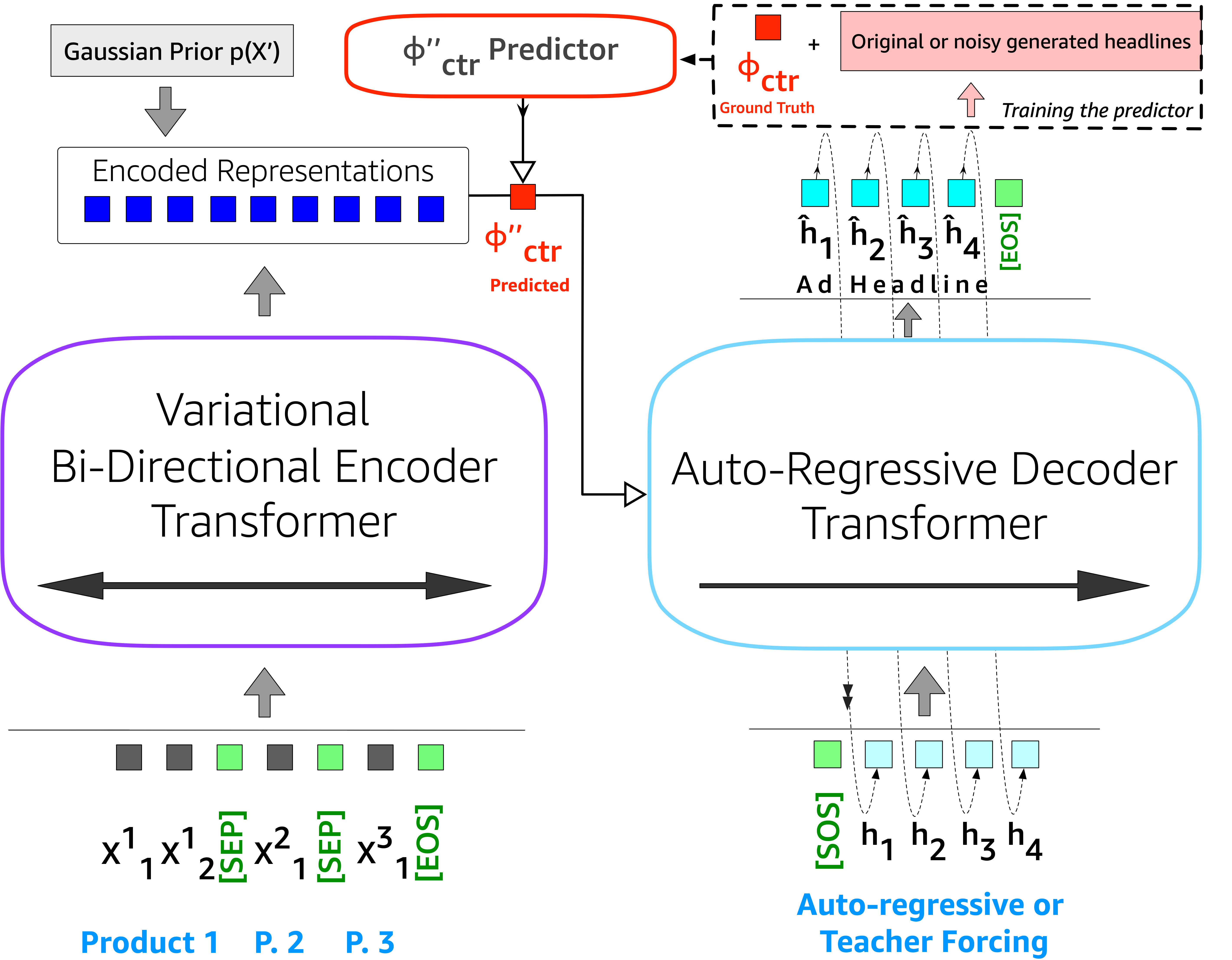}
  \caption{The VBART model conditions the generation on predicted CTR along with
    the encoded representation rather than using control tokens. It adds additional
    discriminator loss using wake-sleep procedure to train the predictor.}
  \label{fig:vbart}
\end{figure}

We also extend previous style-transfer and controllable generation
method~\cite{hu2018toward} that uses variational LSTM auto-encoders. It allows
conditioning of the decoder on a set of structured variables in addition to the
latent output of the encoder as shown in Figure~\ref{fig:vbart}.

We extend the architecture to work with Transformer encoder-decoder. The
pre-trained Transformer encoder probabilistically encodes the input $X$ from
eq.~1 with standard Gaussian prior $p(X')$ to $X'$ such that;
\begin{equation}
  X' \sim q(X'|X)
\end{equation}
Additionally, a Transformer discriminator model is fine-tuned further during
training to predict a CTR estimate $\phi''_{ctr}$. It is trained by pairing the
original CTR $\phi_{ctr}$ with the original human-written headlines and a small
set of noisy generated headlines.

We pose the generation to be conditioned on the estimate $\phi''_{ctr}$ and $X'$
with the distribution:
\begin{equation}
  P_{vt} = \prod_{t=1}^{|\bar{H}|} p(h_t | h_{1:t-1}, X', \phi''_{ctr})
\end{equation}
Following~\cite{hu2018toward}, we update the encoder and the discriminator using
a wake and sleep procedure. The model is updated using the loss:
\begin{equation}
  \begin{split}
    \mathcal{L}_{vt} = {} & \mathbf{KL}(q(X'|X)||p(X'))\\
      & - \lambda_1 \mathbb{E}_{q(X'|X)\phi''_{ctr}}[\log P_{vt}]\\
      & - \lambda_2 \mathbb{E}_{p(X')\phi_{ctr}}[\log \phi''_{ctr}]
  \end{split}
\end{equation}

\clearpage
\section{Experiments}
We primarily experiment with all the discussed methods and baselines to generate
headlines that optimize CTR. We then also experiment to study the effectiveness
of our method to control the length of generation. We finally train a model that
jointly allows to control and optimize both the length and the CTR.

\subsection{Training}
\leadin{Data.} We conducted our experiments on products and human-written
headlines written in English. We used over 500,000 ad campaigns that were created
on Amazon by sellers who have signed-up for advertising. The problem requires 10x
compression of the input and over 50\% abstraction.

We deduplicated all the ads with the same input-output pairs and made sure there
was no overlap in the train, val and test splits. We also made sure that none of
the ads from the val and test sets were used for training of any supporting
models (oracle models, $\phi_{ctr}$ predictors or calculation etc.). We only
selected the ads that comply with ad policies as verified by internal experts.

\leadin{Framework.} We use the HuggingFace~\cite{wolf2020huggingface} pre-trained
versions of the Transformer models. We fine-tune the `large' variants of the
models using Adam~\cite{kingma2017adam} optimizer with early-stopping for upto 15
epochs. For early stopping, we track Rouge-L~\cite{lin2004rouge} on the validation
set with patience of 3 epochs and minimum increment of 0.1. For our VBART
implementation, we refer to the Texar~\cite{hu2019texar} library. We use `fp16'
for faster training and inference latency optimization.

\leadin{COBART.} For training COBART that allows for controlling and optimizing
CTR, we obtain the observed CTR for all the headlines in the training data and
bucketize it into 15 equal sized buckets based on CTR percentile with the prefix
tag `engaging:'. For controlling the length of the headlines, we add a control
tag with three possible values: `short', `medium', and `long' based on whether
the actual headline has $\leq 5$, 6 to 8 or more than 8 words respectively. We use
the prefix `length' for these length control tokens.

We train 3 variants, first only on $\phi_{ctr}$ to compare Rouge-L and CTR
against other methods without effects of user specified length. We then train
COBART only on $\phi_{Length}$ to compare against the baseline length control
method without effects of CTR and finally on both $\phi_{ctr}$ \& $\phi_{Length}$
to study if multiple criteria can be jointly controlled and optimized.

\leadin{SCBART.} For the training of Self-critical variants, $\phi'_{ctr}$ is
estimated using an oracle model that is treated as a black-box and is not updated
during the training of Language Models. We follow the procedure discussed in
Section~\ref{sec:scbart} and experiment with different values of $\lambda$.

\leadin{VBART.} We follow the training procedure indicated in~\cite{hu2018toward}
using Transformer based models. The discriminator is also updated following a
sleep-wake-procedure. Apart from the difference in the backbone LM architecture,
another difference is that we omit the auto-encoder based loss as our model is not
an auto-encoder and thus the generator output cannot be fed back into the encoder.

\subsection{Inference and Deployment}
During inference, the control tokens from training are replaced with the desired
attribute. For instance, $\phi_{Length}$ is replaced with the `short' token if
short headlines are desired, or the token corresponding to the highest CTR bucket
if a completely engaging headline is required. For other models which use
continuous CTR, we use the highest known CTR as the input for all the test
headlines.

We use Beam Search with a beam size of 5, Length Penalty of 1.5 (For $\phi_{ctr}$
experiments), and Repetition Penalty of 2.0 as tuned on the validation set.

All the results are reported for the setting in which the CTR input provided
during inference corresponds to the highest CTR. For instance, if bucket 15 had
the highest CTR during training, then during inference, all headlines are
generated with bucket 15 as the desired CTR.

\subsection{Additional baselines and ablations}
We use the best performing method in~\cite{kanungo2021ad} as our first baseline.
It has already been shown to outperform LSTM and RNN based methods and we thus
exclude them.

For a more thorough comparison, we also fine-tuned the large variants of
T5~\cite{raffel2020exploring} and ProphetNet~\cite{qi2020prophetnet} to generate
the headlines.

As another baseline, we fine-tune the BART model using historical ads that have
observed CTR $\phi_{ctr}$ higher than the median CTR of the entire training data.
The idea being that the model should learn to generate headlines with only the
style of high CTR headlines.

To incorporate continuous $\phi_{ctr}$, we bucketize the feature into multiple
percentile based buckets. We study the effect of using fine-grained buckets by
using different number of buckets in the VBART and COBART models.

In the COBART model, we propose that adding the control tokens as an input to the
encoder improves performance. We thus also study the effect of concatenating the
$\phi_{ctr}$ value to the output of the encoder rather than as input.

We also experiment with different values of $\lambda$ in the total Self-critical
loss function shown in equation 7.

Finally, to study the impact of length control, we compare our COBART variant
that is trained to control generation length against our BART model whose length
is controlled using Length Penalty~\cite{murray2018correcting}. We try to vary the
penalty value to obtain desired lengths.

\section{Results}
\begin{figure}[t]
  \centering
  \includegraphics[width=0.9\columnwidth]{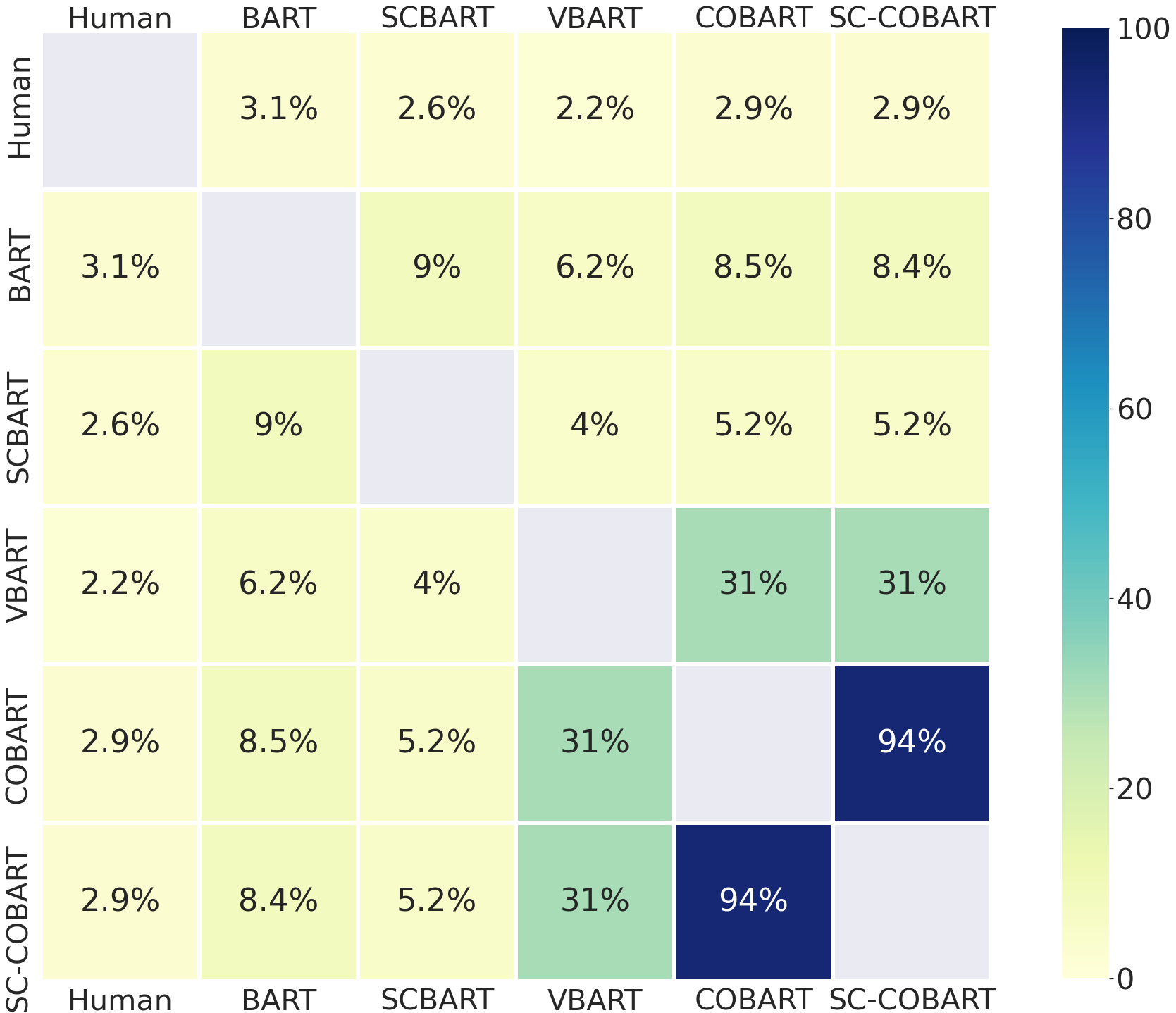}
  \caption{The \% of generated headlines that are exactly the same across
    different sources.}
  \label{fig:overlap}
\end{figure}

We first evaluate if the use of different techniques with the same pre-trained
model and the same training data results in different generated headlines.
Figure~\ref{fig:overlap} illustrates that all the three methods generate more
than 90\% new headlines compared to baseline and human-written reference. There
is a very high overlap between COBART and SC-COBART.

\begin{table*}[t]
  \begin{tabular}{@{}llccc@{}}
    \toprule
    Model & Inputs & Rouge-L & CTR v. Human & CTR v. Baseline\\
    \midrule
    \multicolumn{5}{@{}l}{\textit{Transformer Variants and Baselines}}\\
    \quad UniLM SCST with Rouge-L~\cite{dong2019unified,kanungo2021ad} & Titles & - & 7.84\% & -\\
    \quad T5~\cite{raffel2020exploring} & Titles & -0.72\% & -1.96\% & -9.09\%\\
    \quad ProphetNet~\cite{qi2020prophetnet} & Titles & 1.57\% & 2.16\% & -5.27\%\\
    \quad BART~\cite{lewis2019bart} & Titles & 10.75\% & -3.73\% & -10.73\%\\
    \quad BART trained on filtered high CTR data & Titles, $\phi_{ctr}$* & 5.18\% & 4.90\% & -2.73\%\\
    \midrule
    \multicolumn{5}{@{}l}{\textit{SC-BART and ablations}}\\
    \quad SC-BART : $\lambda = 0.0$ & Titles, $\phi'_{ctr}$* & -23.33\% & -10.20\% & -16.73\%\\
    \quad SC-BART : $\lambda = 0.1$ & Titles, $\phi'_{ctr}$* & -17.43\% & -5.88\% & -12.73\%\\
    \quad SC-BART : $\lambda = 0.5$ & Titles, $\phi'_{ctr}$* & 16.87\% & 7.25\% & -0.55\%\\
    \quad SC-BART : $\lambda = 0.9$ & Titles, $\phi'_{ctr}$* & 18.94\% & 3.92\% & -3.64\%\\
    \midrule
    \multicolumn{5}{@{}l}{\textit{Variational BART and ablations}}\\
    \quad VBART : 2 $\phi''_{ctr}$ buckets & Titles, $\phi_{ctr}$*, $\phi''_{ctr}$ & 11.40\% & 3.92\% & -3.64\%\\
    \quad VBART : 15 $\phi''_{ctr}$ buckets & Titles, $\phi_{ctr}$*, $\phi''_{ctr}$ & 14.02\% & 6.27\% & -1.45\%\\
    \quad VBART : Continuous $\phi''_{ctr}$ & Titles, $\phi_{ctr}$*, $\phi''_{ctr}$ & 19.95\% & 10.39\% & 2.36\%\\
    \midrule
    \multicolumn{5}{@{}l}{\textit{COBART for $\phi_{ctr}$ and ablations}}\\
    \quad Concatenating $\phi_{ctr}$ with encoder embeddings & Titles, $\phi_{ctr}$ & 20.58\% & 11.37\% & 3.27\%\\
    \quad COBART : 2 $\phi_{ctr}$ buckets & Titles, $\phi_{ctr}$ & 6.91\% & 0.15\% & -7.13\%\\
    \quad \textbf{COBART : 15 $\phi_{ctr}$ buckets} & Titles, $\phi_{ctr}$ & \textbf{21.89\%} & \textbf{11.76\%} & \textbf{3.64\%}\\
    \quad \textbf{SC-COBART} & Titles, $\phi_{ctr}$, $\phi'_{ctr}$* & \textbf{25.82\%} & \textbf{14.12\%} & \textbf{5.82\%}\\
    \bottomrule
  \end{tabular}
  \caption{The \% improvement over UniLM baseline in overlap with ad policy
    compliant human-written headlines (Rouge-L) and the estimated CTR. We also
    report \% improvement in the estimated CTR compared to human-written headlines.
    The COBART model that optimizes $\phi_{ctr}$ outperforms other techniques. It
    is further boosted when combined with SCST. (*) indicates the feature is only
    used during training. Other features are replaced with the highest value
    during inference.}
  \label{tab:main}
\end{table*}

\subsection{Overlap with human-written headlines}
We measure several overlap metrics of the generated model headlines with the
reference human-written headlines that were approved by internal experts. We
report the Rouge-L~\cite{lin2004rouge} F1 metric. All the other commonly used
overlap metrics such as CIDEr, BLEU-4, METEOR etc. were in complete agreement
with Rouge-L and we thus omit them.

Table~\ref{tab:main} shows the Rouge-L metrics across all the experiments.

In agreement with previously published results which indicate that BART performs
well on summarization related tasks~\cite{lewis2019bart}, we observe that BART
achieves better Rouge-L score than T5, ProphetNet, and UniLM that is fine-tuned
with SCST for Rouge-L. It shows an improvement of 10.75\% in the Rouge-L score
over UniLM. In the case when the BART model is only fine-tuned on the top-50
percentile of data that has the highest CTR, we observe that the Rouge-L score
drops significantly from an improvement of 10.75\% to an improvement of 5.18\%.
This could be an indication that the model is not able to generalize to all kinds
of products and brands present in the complete dataset and learn commonly
associated product features.

On using the Self-critical framework to optimize CTR for the BART model, we see
an improvement in Rouge-L for higher values of $\lambda$. On increasing the
contribution of the CTR based loss (lower $\lambda$ values), we see that the CTR
improves at the expense of Rouge-L. Moreover, model does not learn when only the
CTR based loss is used.

In the case of Variational BART generator, we see that the Rouge-L improves
consistently across all the experiments. We see the highest improvement when the
CTR is used directly as a continuous value which shows that dividing CTR into
categorical buckets does lead to loss of information.

Similar to Variational BART, adding finer and more CTR buckets improves the
performance of the COBART model and yields the highest Rouge-L score of any model.
Increasing the number of buckets beyond 15 does not yield any benefit. On
comparing the results to an ablation in which the CTR is concatenated with output
Key and Value embeddings from the encoder, we see that the performance decreases
compared to when it is given as input to the encoder. Combining the Self-critical
framework and COBART model boosts and yields the overall highest Rouge-L score.

\subsection{Estimated CTR}
We use the oracle CTR model to estimate the CTR of the generated headlines. In
Table~\ref{tab:main}, we also report the \% improvement in the estimated CTR
compared to the estimated CTR of the 1) Baseline Self-critical UniLM model 2)
Human-written headlines.

\leadin{Trade-off between Rouge-L and CTR.} In some interesting scenarios the CTR
improved but the Rouge-L score did not and vice-versa (UniLM, T5 and ProphetNet v.
BART etc.). We observe that at times the models may generate more novel words
(adjectives etc.) that improve CTR but those words do not align with the reference
headline and specific features of the products and brands. There is thus a
trade-off in such scenarios and in the ideal scenario, both the overlap scores and
CTR should increase.

The best SC-BART variant outperforms BART trained without SCST in terms of CTR. It
also nearly equals the CTR of the baseline SC-UniLM model at a significantly higher
Rouge-L score. For VBART, using continuous $\phi''_{ctr}$ values as input to the
generator gives the highest boost to the estimated test CTR. The model outperforms
the baseline model by 2.36\% and also outperforms the human-written headlines by
10.39\%. Using bucketized CTR in this setting leads to worsened performance.

On using just 2 buckets for CTR with the COBART model, we see a big drop in the
estimated test CTR compared to the model that uses 15 buckets. This shows that a
certain level of hyper-parameter tuning is required for continuous control
characteristics. On comparing the results to an ablation when the CTR is
concatenated with encoder output, we see that the performance decreases compared
to when it is given as input to the encoder. This highlights the advantage of
conditioning the encoder output on the control tokens. COBART model outperforms
baseline CTR by 3.54\% and human-written headline CTR by 11.76\%. When combined
with SCST, we see a further boost in the CTR performance even when only 6\% of all
the test headlines change.

\subsection{Controlling Length}
Table~\ref{tab:length} shows the median length of different types of headlines
generated using the BART + Length Penalty baseline and the COBART variant trained
to control length. We see that COBART is effectively able to control the length of
the generated output whereas BART + Length Penalty hits an upper bound. On
increasing the penalty beyond 2.0 (We tried upto 10) we see that the effective
length actually decreases due to diminishing relative effects of the penalty.

81\% of long and medium \& 35\% of short and medium headlines generated by the
baseline BART were exactly the same. Whereas, only 4\% of long and medium and fewer
than 1\% of short and medium headlines were same in COBART.

Thus, our results show that COBART is a much more reliable method to control
length. It is able to generalize to all kinds of products without needing parallel
training corpora with different headlines available for the same product.

\begin{table}[t]
  \begin{tabular}{@{}lcc@{}}
    \toprule
    Control Tag & \begin{tabular}{@{}c@{}}Average\\num. of\\characters\end{tabular} & \begin{tabular}{@{}c@{}}Average\\num. of\\words\end{tabular}\\
    \midrule
    BART - Length Penalty: 0.25 & 31 & 5\\
    BART - Length Penalty: 1 & 37 & 6\\
    BART - Length Penalty: 1.5 & 40 & 6\\
    BART - Length Penalty: 2.0 & 35 & 6\\
    COBART - $\phi_{Length}$ : Short & 30 & 5\\
    COBART - $\phi_{Length}$ : Medium & 42 & 7\\
    COBART - $\phi_{Length}$ : Long & 47 & 9\\
    \bottomrule
  \end{tabular}
  \caption{Controlling headline length using COBART $\phi_{Length}$ and Length
    Penalty (BART) during inference}
  \label{tab:length}
\end{table}

\section{Qualitative Analysis and Discussion}

\subsection{Optimizing CTR and Rouge-L}
Table~\ref{tab:qual} illustrates model generated headlines for a set of products
from the test set. Across these and other samples in the test set, we notice that
the models with higher CTR tend to generate headlines that are more descriptive
and complete and have much rarer keyword stuffing issues. On comparing to the set
of human-written headlines, we can see that the models are able to generate
headlines with great quality. All the models are also able to effectively combine
information from all the products in the ad. We also observe that even though the
train and test splits do not have any product-title overlap, the models are able
to learn phrases associated with certain brands and use them for new headlines
generated for their newer products. This behavior is reflected in the increment in
the Rouge-L score.

We also analyzed if there are any particular words that are used more often by one
model over another. We thus compared the frequency of words across all headlines
generated by different models. It is interesting to note that compared to BART, the
SC-COBART model uses `Adventure' 9x times, `Brighten' 8x times, `Always' 7x times,
`Everyday' 7x times and `Powerful' 4x times. At the same time it uses `Good' 7
times fewer, `24/7' 5.5 times fewer, `favorite' 3 times fewer and `Pain' 2 times
fewer. The increased CTR and quality cannot be attributed to just the presence of
these words but it does show a trend that the proposed model is able to use more
attractive words that entice higher CTR.

\begin{table*}[t!]
  \footnotesize
  \setlength{\tabcolsep}{3pt}
  \renewcommand{\arraystretch}{1.45}
  \begin{tabular}{@{}p{2.05cm}p{2.05cm}>{\itshape}p{2.05cm}>{\itshape}p{2.05cm}>{\itshape}p{2.05cm}>{\itshape}p{2.05cm}>{\itshape}p{2.05cm}>{\itshape}p{2.05cm}@{}}
    \toprule
    Product Title 1 & Product Title 2 & Human & BART & SCBART & VBART & COBART $\phi_{ctr}$ & SC-COBART $\phi_{ctr}$\\
    \midrule
    2 Pack 12 Colors Makeup Naked Eyeshadow \ldots & 60 Colors Eyeshadow Palette, 4 in1 Color Board \ldots & Vibrant Colors, Most Definitely Worth Eyeshadow & Best Eyeshadow Palette Set & BestLand Eyeshadow palette & Best Best Gift for Your Loved Ones & Shimmering Glitter Eyeshadow Palette & Shimmering Glitter Eyeshadow Palette Set\\
    \addlinespace
    {[2020 Upgraded Version]} ZeeHoo Wireless Car Charger,15W Qi \ldots & ZeeHoo Wireless Car Charger,10W Qi Fast Charging Auto-Clamping \ldots & Fast Charging Auto Clamping Wireless Car Charger & ZeeHoo Wireless Car Charger Mount & ZeeHoo Wireless Car Charger & Auto Clamping Wireless Car Charger Mount & Auto Clamping Wireless Car Charger Mount by ZeeHoo & Auto Clamping Wireless Fast Charging Car Mount\\
    \addlinespace
    New! - Shush Biker - High Performance Hearing \ldots & NEW - Shush Acoustic - Universal-Fit \ldots & New: High Fidelity Earplugs with Ceramic Filter & Earplugs with Ceramic Filter - Shush & Shush Earplugs - Hear What You've & Earplugs with Ceramic Filter - Superior Sound & Earplugs with Ceramic Filter for Live Music & Earplugs with Ceramic Filter for Motorcyclists\\
    \addlinespace
    Teamoy Sewing Machine Case, Travel Tote Bag \ldots & Teamoy Sewing Machine Carrying Case, Sewing \ldots & Sewing Machine Carry Made Easy & Protect Your Sewing Machine in Style & Sewing Machine Tote Bag for Home Sewing & Easy to Carry Your Sewing Machine & Carry your sewing machine wherever you go & Carry your sewing machine wherever you go\\
    \addlinespace
    Happyluxe Travel Pillow | Small Pillow for Neck \ldots & Happyluxe Travel Pillow | Small Pillow for Neck \ldots & The Perfect Pillow for Travel and Lounging & The Perfect Pillow for Neck Pain Relief & The Perfect Little Pillow For Your Little One & The Perfect American Made Travel Pillow for Camping & The Perfect American Made Travel Pillow for You & The Perfect American Made Travel Pillow for You\\
    \addlinespace
    Tria Age-Defying Smooth Beauty Laser \ldots & Tria Beauty Age-Defying Eye Wrinkle \ldots & Anti Wrinkle, Anti Aging FDA Cleared Laser & Wrinkle Correcting Laser for Younger Skin & Age defying skincare that doesn't look expensive & Age Defying Beauty Laser That Actually Works & Age defying treatments that actually work & Age defying skincare laser technology\\
    \addlinespace
    TRX PRO3 Suspension Trainer System Design \& Durability| \ldots & TRX GO Suspension Trainer System: Lightweight \ldots & Full body training at home, outdoors or on the go & Portable full body workout at home or outdoors & Get a Full Body Workout At Home or On & Get the ultimate go-anywhere gym & Get Ready For Summer? Get Your Suspension System & Get Ready To Get Your Suspension Starter Kit\\
    \bottomrule
  \end{tabular}
  \caption{Generated headlines across different models for multiple input products
    (First 2 of all products shown)}
  \label{tab:qual}
\end{table*}

\begin{table*}[t!]
  \footnotesize
  \setlength{\tabcolsep}{3pt}
  \renewcommand{\arraystretch}{1.45}
  \begin{tabular}{@{}p{2.05cm}p{2.05cm}>{\itshape}p{2.05cm}>{\itshape}p{2.05cm}>{\itshape}p{2.05cm}>{\itshape}p{2.05cm}>{\itshape}p{2.05cm}>{\itshape}p{2.05cm}@{}}
    \toprule
    Product Title 1 & Product Title 2 & \begin{tabular}{@{}l@{}}BART - LP 0.25\\(Short)\end{tabular} & \begin{tabular}{@{}l@{}}BART - LP 1\\(Medium)\end{tabular} & \begin{tabular}{@{}l@{}}BART - LP 1.5\\(Long)\end{tabular} & \begin{tabular}{@{}l@{}}COBART $\phi_{Length}$\\(Short)\end{tabular} & \begin{tabular}{@{}l@{}}COBART $\phi_{Length}$\\(Medium)\end{tabular} & \begin{tabular}{@{}l@{}}COBART $\phi_{Length}$\\(Long)\end{tabular}\\
    \midrule
    Kids Easy Lazy Halloween Costume Shirt Skeleton \ldots & Womens Mens Easy Lazy Halloween Costume Shirt \ldots & Fun Easy Lazy Halloween Shirts & Easy Lazy Halloween Costume Shirts & Easy Lazy Halloween Costume Shirts & Fun Halloween costume shirts & Fun Halloween shirts for the whole family & Fun easy lazy costume shirts for kids and adults\\
    \addlinespace
    Thompson's TH.010502-18 Waterseal Fabric Seal - Aersol & Thompson's TH.087731-42 WaterSeal Oxy Foaming \ldots & Shop Thompson's WaterSeal Products & Shop Thompson's WaterSeal Products Today & Shop Thompson's WaterSeal Products Today & Protect Your Surfaces \& Furniture & Protect Your Surfaces \& Furniture from Water Damage & Protect Your Surfaces \& Furniture From Water \& Dirt\\
    \addlinespace
    Capsuline Clear Gelatin Empty Capsules 000 1000 \ldots & Capsuline Colored Gelatin Empty Capsules Size 0 Red \ldots & Capsuline Colored Gelatin Capsules & Make Your Own Supplements with Capsuline Gelatin & Make Your Own Supplements with Capsuline Gelatin & Make Your Own Supplements & Make Your Own Supplements with Capsuline Capsules & Make your own Supplements at home with Capsuline\\
    \addlinespace
    Newport Vessels 8-Feet 10-Inch Dana Inflatable Sport \ldots & Newport Vessels 9-Feet 6-Inch Del Mar Inflatable Sport \ldots & Inflatable Sport Tender Dinghy Boats & Inflatable Sport Tender Dinghy Boats & Inflatable Sport Tender Dinghy Boats & Sport Tender Dinghy Boats & Sport Tender Dinghy Boats from Newport Vessels & Sport Tender Dinghy Boats - Fun for All Ages\\
    \bottomrule
  \end{tabular}
  \caption{Generation with different lengths using Length Penalty (LP) with BART
    and COBART (First 2 of all products shown)}
  \label{tab:quallength}
\end{table*}

\subsection{Controlling Length}
Table~\ref{tab:quallength} illustrates headlines generated by the baseline approach
and our proposed approach to control length. It is evident that COBART is able to
generate headlines with different lengths much more consistently. The short
headlines tend to describe the products in a few words, medium headlines add some
additional words and the long headlines are consistently most descriptive.

On examining the model that controls both CTR and length, we see that the model is
able to jointly control both the parameters effectively and yields results
consistent with other experiments. For instance, on examining the subset of test
ads with short human written headlines, we see that COBART is also able to generate
short headlines 99.83\% times while still yielding CTR improvement of 6.7\%.
Table~\ref{tab:teaser} illustrates this joint control over both the length and CTR
(3 chosen buckets). This enables automated ad headline generation that works across
all ad formats, requirements, and also optimizes CTR using a single model.

\section{Conclusion}
We propose a novel solution to the challenging and high-impact problem of ad
headline generation. Our proposed method is able to control and optimize ad
headline generation by using control tokens, SCST and the BART model. We compare
our model to strong baselines and ablations and demonstrate its efficacy both
quantitatively and qualitatively. We would continue to evaluate the model behavior
over an extended period and experiment with more characteristics control tokens.

\begin{acks}
Thanks to Ramaiah M S and Manisha Verma for proof-reading and scrutinizing the
paper.
\end{acks}

\clearpage

\nocite{awsneuron,xu2021vae}
\bibliographystyle{ACM-Reference-Format}
\bibliography{references}

\appendix
\onecolumn
\section{Reproducing Experiments on Your Data}
\begin{enumerate}
  \item Process your data and convert it to the HuggingFace Dataset format (Link).
  \item Compute observed CTR, length or any other desired control tag and
    obtain/train an oracle model if needed
  \item Setup Summarization / Generation script that uses HuggingFace Transformers
    (Link)
    \begin{enumerate}
      \item BART (Link)
      \item T5 (Link)
      \item ProphetNet (Link)
      \item UniLM (Link)
    \end{enumerate}
  \item COBART
    \begin{enumerate}
      \item Add control tag prefixes to the input either as known tokens or
        extended tokens
      \item Use the Summarization / Generation script for training
    \end{enumerate}
  \item SCBART
    \begin{enumerate}
      \item Calculate the reward function using the oracle model (Link)
      \item Compute $\mathcal{L}_{RL}$ using headline generated using sampling.
      \item Use convex combination of $\mathcal{L}_{RL}$ and $\mathcal{L}_{BART}$
      \item Update the parameters
    \end{enumerate}
  \item VBART
    \begin{enumerate}
      \item Refer to the VAE auto-encoder implementation (Link)
      \item Encode the input to X' and decode to generate the headline
      \item Use the loss without the auto-encoder component
    \end{enumerate}
  \item SC-COBART
    \begin{enumerate}
      \item Follow COBART steps
      \item Follow SCBART steps to use convex combination of $\mathcal{L}_{RL}$
        and $\mathcal{L}_{COBART}$
    \end{enumerate}
\end{enumerate}

\end{document}